\documentclass{article}
\usepackage{titlesec}
\usepackage{graphicx}
\usepackage{arxiv}
\usepackage[utf8]{inputenc}
\usepackage{setspace}
\usepackage{geometry}
\usepackage{indentfirst}
\usepackage{appendix}
\usepackage{amsmath} 
\usepackage{amssymb} 

\usepackage{hanging}
\newcommand{\hangindentlength}{0.5in} 

\usepackage[T1]{fontenc}    
\usepackage{hyperref}       
\usepackage{url}            
\usepackage{booktabs}       
\usepackage{amsfonts}       
\usepackage{nicefrac}       
\usepackage{microtype}      
\usepackage{lipsum}		
\usepackage{graphicx}
\usepackage{doi}
\usepackage{tabularx}
\usepackage{booktabs}
\usepackage{threeparttable}

\title{Achieving Semantic Consistency: Contextualized Word Representations for Political Text Analysis}

\date{December 3, 2024}	

\author{ \href{https://orcid.org/0000-0002-0883-4574}{\includegraphics[scale=0.06]{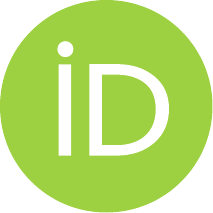}\hspace{1mm}Ruiyu Zhang} \\
	Department of Applied Social Sciences\\
    The Hong Kong Polytechnic University\\
	ruiyu.zhang@polyu.edu.hk\\
	\And
	\href{https://orcid.org/0000-0002-0275-117X}{\includegraphics[scale=0.06]{orcid.pdf}\hspace{1mm}Lin Nie} \\
	Department of Applied Social Sciences\\
    The Hong Kong Polytechnic University\\
	lin-apss.nie@polyu.edu.hk\\
	\And
        \href{ }
	{\hspace{1mm}Ce Zhao} \\
	School of Science\\
    Harbin Institute of Technology (Shenzhen)\\
        \And
        \href{ }
	{\hspace{1mm}Qingyang Chen} \\
	Department of Applied Social Sciences\\
    The Hong Kong Polytechnic University\\
}

\renewcommand{\headeright}{Preprint}
\renewcommand{\undertitle}{Preprint}
\renewcommand{\shorttitle}{Achieving Semantic Consistency}

\hypersetup{
pdftitle={Achieving Semantic Consistency Zhange et al},
pdfsubject={cs.CL, econ.GN},
pdfauthor={Ruiyu Zhang, Lin Nie, Ce Zhao, Qingyang Chen},
pdfkeywords={Contextual Word Embeddings, Text Analysis, Natural Language Processing,  Semantic Shift},
}

\begin{document}
\maketitle

\begin{abstract}
Accurately interpreting words is vital in political science text analysis; some tasks require assuming semantic stability, while others aim to trace semantic shifts. Traditional static embeddings, like Word2Vec effectively capture long-term semantic changes but often lack stability in short-term contexts due to embedding fluctuations caused by unbalanced training data. BERT, which features transformer-based architecture and contextual embeddings, offers greater semantic consistency, making it suitable for analyses in which stability is crucial. This study compares Word2Vec and BERT using 20 years of People’s Daily articles to evaluate their performance in semantic representations across different timeframes. The results indicate that BERT outperforms Word2Vec in maintaining semantic stability and still recognizes subtle semantic variations. These findings support BERT’s use in text analysis tasks that require stability, where semantic changes are not assumed, offering a more reliable foundation than static alternatives.
\end{abstract}


\section{Introduction}
Robustly interpreting word meanings over time is crucial for understanding socio-political and cultural dynamics. Political science scholars often rely on text-based analyses to gauge how concepts evolve within media and policy discourse. It is particularly important to maintain stable semantic interpretations when examining long-term shifts in stereotypes, party alignments, or public sentiments (Rodman, 2020; Rodriguez \& Spirling, 2022). However, ensuring that our models capture authentic semantic transformations, rather than artifacts of data imbalance or noise, remains a fundamental challenge in natural language processing (NLP). 

Word2Vec (Mikolov et al., 2013) introduced shallow neural networks to produce distributed word representations in a high-dimensional vector space, which represented a major milestone in NLP. Word2Vec’s two primary architectures—Continuous Bag of Words (CBOW) and Skip-gram—enabled the effective learning of co-occurrence patterns. CBOW predicts a target word from its surrounding context, optimizing computational speed, while Skip-gram predicts surrounding words from a target word and tends to perform better on smaller datasets. Due to its ability to capture semantic associations, Word2Vec has been employed in political science: it has helped identify electoral violence (Muchlinski et al., 2021), measure opposing political terms to assess polarization (Pereira et al., 2024), and examine how exposure to celebrities from stigmatized groups might reduce prejudice (Alrababa’h et al., 2021). Scholars have also used Word2Vec to expand specialized dictionaries for political event coding (Radford, 2021).

Other static embedding approaches, including GloVe (Pennington et al., 2014) and FastText (Bojanowski et al., 2017), similarly revolutionized text analysis. The former derives embeddings by leveraging global word co-occurrence statistics, while the latter extends Word2Vec by incorporating subword character n-grams, effectively handling out-of-vocabulary words. These static techniques have proven instrumental for analyzing broad, long-term semantic shifts, such as tracking changes in gender and ethnic stereotypes over a century (Garg et al., 2018) or examining cultural transformations related to “class” and “power” (Kozlowski et al., 2019). Yet, static embeddings assign a single vector to each word irrespective of its context, risking instability and ambiguity, especially when training data are partitioned into narrow temporal slices. In short-term analyses, unbalanced training samples can artificially inflate or obscure semantic changes and generate misleading interpretations.

To address these limitations, transformer-based architectures—especially BERT (Devlin et al., 2018)—introduced contextual word embeddings that adapt to each token’s specific linguistic environment. BERT’s true bidirectionality hinges on the self-attention mechanism (Vaswani et al., 2017), which enables it to capture both local and global dependencies in a sequence. Beyond better handling polysemy, BERT typically produces more stable representations for contexts in which minimal semantic drift should be assumed. This stability is beneficial for a variety of NLP tasks, such as text classification and annotating (Rogers \& Rumshisky, 2021; Yang et al., 2022). Fine-tuned or extended BERT models have recently been applied to sentiment and stance classification (Bestvater and Monroe, 2023; Widmann and Wich, 2023; Burnham, 2024) as well as topic and content classification (Wang, 2023; Müller and Fujimura, 2024).

Few studies have offered a direct empirical analysis of how static and contextual models perform compared to varying temporal intervals—particularly in short spans, where semantic drift should be minimal. The present study seeks to fill this gap by comparing the ability of Word2Vec and BERT to provide stable word interpretations and detect authentic semantic changes. Using 20 years of People’s Daily articles, we train each model at multiple resolutions (20, 10, 5, and 3 years). We then measure model performance using several metrics, including Semantic Displacement (SD), Mean Temporal Similarity (MTS), Rate of Semantic Change (RSC), and Local Neighborhood Stability (LNS). We expect BERT to offer greater stability and more effectively differentiate true semantic evolution from fluctuations caused by uneven data, thereby avoiding spurious signals.

This research has practical implications for political science, media studies, and policy analysis. Scholars frequently rely on textual data to trace how narratives around public policy, social movements, or international relations emerge and shift over time. In contexts where year-to-year changes in discourse might be subtle, misclassifying random noise as genuine drift poses significant risks to inference. By empirically documenting BERT’s advantages in stable embedding generation, we aim to bolster the methodological foundations for investigating political language. Our findings underscore the importance of contextual embeddings for ensuring consistent, consecutive interpretations of words and concepts in large-scale longitudinal studies, ultimately enriching analyses of socio-political phenomena.

\section{Methodology}
The dataset used for this study comprises articles from the People’s Daily, a prominent Chinese newspaper, from 2004 to 2023. This extensive dataset captures changes in the country’s political, economic, cultural, and social discourse. It offers a rich linguistic resource for studying how language evolves during major events and policy changes. The dataset was preprocessed, tokenized by jieba, and saved in yearly segments to facilitate efficient model training and analysis. To enable detailed analysis, it was segmented into four time frames: 20 years (2004–2023, 389.6M tokens, 654.1K articles), 10 years (2014–2023, 212.9M tokens, 321.5K articles), 5 years (2019–2023, 94.4M tokens, 129K articles), and 3 years (2021–2023, 53.9M tokens, 71K articles). Examining these different time frames allows us to assess the stability of word embeddings and analyze the varying dynamics of semantic representations. The longer time spans (20 and 10 years) permit examination of broader chronological linguistic changes, while the shorter periods (5 and 3 years) provide insight into more immediate shifts and fluctuations in meaning.

The study’s primary aim is to compare the performance of BERT and Word2Vec in capturing semantic changes during different time frames (Figure 1 visualizes the assessment procedure for model training and comparison). Word2Vec models were trained for each year using Skip-gram architectures. The Skip-gram model was chosen to emphasize the learning of rare word representations, given its ability to capture occasional semantic shifts in vocabulary. Word2Vec models for each year were then aligned with a common vector space using Orthogonal Procrustes analysis (Yao et al., 2018; Rodman, 2020). This alignment was critical to ensure consistency when comparing embeddings over time; it allows us to determine whether word meanings shifted or remained stable across different periods. The alignment step employed shared vocabulary between models to compute an orthogonal transformation matrix, aligning each target model to the base model (from 2004). For BERT embeddings, we utilized the pre-trained BERT-base-Chinese model to generate contextual embeddings for our filtered newspaper data. For each sentence that contained at leaset one of our target keywords (see Appendix 1), we extracted the 768-dimensional embedding for each token from BERT’s last hidden layer. Unlike Word2Vec, BERT’s contextual embeddings do not require explicit retraining for different time periods since they inherently capture context-specific meanings. We processed the texts in batches of 128 sequences (each sequence was truncated to a maximum length of 512 tokens) and computed averaged embeddings for each keyword across all of its occurrences in the texts. This approach generates more stable and consistent word representation over time while capturing the nuanced usage of keywords in Chinese-language contexts.

After obtaining the Word2Vec and BERT embeddings, we performed z-score standardization on all word vectors to ensure comparability across embedding spaces. For each model year, we standardized the embeddings by subtracting the mean and dividing by the standard deviation computed across all words in the vocabulary, transforming the vectors to have zero mean and unit variance. While this approach normalizes the scale, we verified that it does not distort genuine semantic relationships across years. This standardization step was crucial for making fair comparisons between different embedding models and across multiple time periods, as it normalizes the scale of the features while preserving the relative relationships between words.

\textbf{Figure 1. Assessment Procedure}

\begin{figure}[h!]
    \centering
    \includegraphics[width=\textwidth]{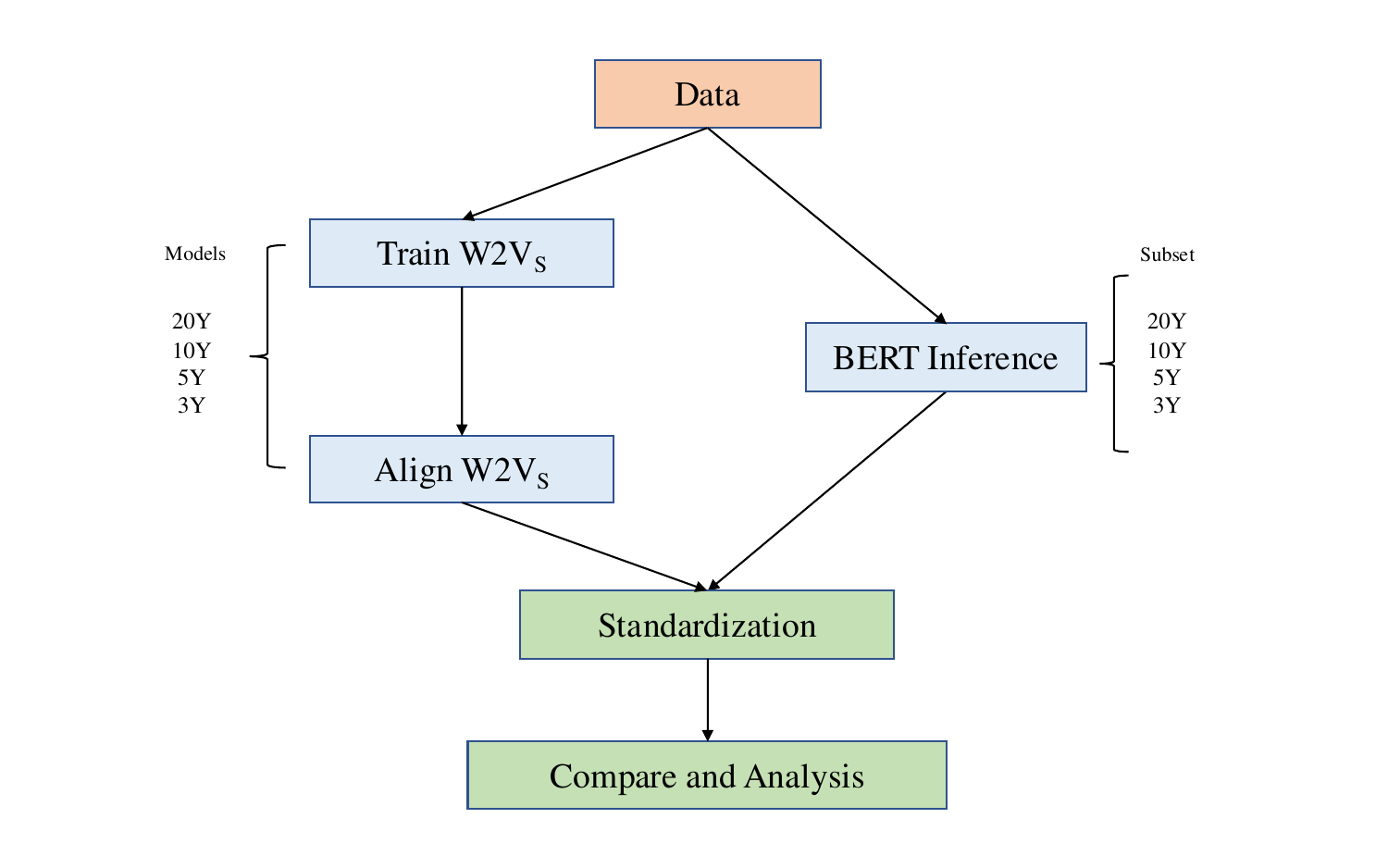} 
\end{figure}

We employed four metrics to evaluate the ability of BERT and Word2Vec models to capture semantic changes and ensure stability. The remainder of the section describes each in turn.

\vskip 0.2 in

\textbf{\textit{Semantic Displacement}}

SD measures the extent to which a word’s meaning shifted from the beginning to the end of a given time period. It calculates the cosine similarity between a word’s embedding in the initial year (e.g., 2004) and its embedding in the final year (e.g., 2023). This metric is crucial for understanding long-term semantic shifts, especially in political science, where shifts in terminology reflect broader sociocultural transformations. The SD for word w, SD(w), from the initial year to the final year is defined as:

\vskip 0.2 in

\[
SD(w) = \cos(v_{t_1}, v_{t_N})
\]

\vskip 0.2 in

Where:
\begin{itemize}
    \item \( v_{t_1} \) is the word embedding in the initial year.
    \item \( v_{t_N} \) is the word embedding in the final year.
    \item \( \cos(v_{t_1}, v_{t_N}) \) represents the cosine similarity between the word embeddings at the initial and final years.
\end{itemize}

\vskip 0.2 in

\textbf{\textit{Mean Temporal Similarity}}

MTS quantifies the average cosine similarity between a word’s vector representation across consecutive time frames. This metric provides insight into the degree of semantic drift, and helps determine whether word meanings evolve gradually or shift abruptly. For instance, words related to technology or societal changes might have a high temporal distance due to rapid shifts in their contextual usage, reflecting evolving discourse. The MTS for word w, MTS(w), across consecutive years is defined as:

\vskip 0.2 in

\[
MTS(w) = \frac{1}{N-1} \sum_{i=1}^{N-1} \cos(v_{t_i}, v_{t_{i+1}})
\]

\vskip 0.2 in

Where:
\begin{itemize}
    \item \( \cos(v_{t_i}, v_{t_{i+1}}) \) represents the cosine similarity between word embeddings at consecutive years \( t_i \) and \( t_{i+1} \).
    \item \( N \) is the total number of time frames.
\end{itemize}

\vskip 0.2 in

\textbf{\textit{Rate of Semantic Change}}

The RSC measures how quickly a word’s meaning evolves during different time frames. It computes the average change in cosine similarity between a word’s vector in consecutive years. A lower cosine similarity between years suggests a faster rate of semantic change, indicating that the word’s meaning is shifting significantly. Words linked to political events or emerging technologies often have higher rates of semantic change, reflecting dynamic discourse in response to societal developments. The RSC for word w, RSC(w), across consecutive years is defined as:

\vskip 0.2 in

\[
RSC(w) = \frac{1}{N-1} \sum_{i=1}^{N-1} \left( 1 - \cos(v_{t_i}, v_{t_{i+1}}) \right)
\]

\vskip 0.2 in

Where:
\begin{itemize}
    \item \( \cos(v_{t_i}, v_{t_{i+1}}) \) represents the cosine similarity between word vectors at consecutive years \( t_i \) and \( t_{i+1} \).
    \item \( N \) is the total number of time frames.
\end{itemize}

\vskip 0.2 in

\textbf{\textit{Local Neighborhood Stability}}

LNS assesses the consistency of the local context of a word over different time periods. It measures the similarity of a word’s nearest neighbors in the embedding space across two time frames. Using top-k nearest neighbors (k=10), this metric evaluates the degree of overlap between neighbors in different years. The LNS (w, t1, t2) for word w between times t1 and t2 is:

\vskip 0.2 in

\[
LNS(w, t_1, t_2) = \frac{| N_k(w, t_1) \cap N_k(w, t_2) |}{k}
\]
\vskip 0.2 in

Where:
\begin{itemize}
    \item \( N_k(w, t) \) represents the set of \( k \) nearest neighbors of word \( w \) at time \( t \).
    \item The \( k \) nearest neighbors for a word \( w \) are determined by ranking all other words in the vocabulary based on their cosine similarity to \( w \) at time \( t \).
    \item The top \( k \) words with the highest cosine similarity are selected as the nearest neighbors, excluding the word \( w \) itself.
    \item \( | A \cap B | \) denotes the cardinality of the intersection of sets \( A \) and \( B \).
\end{itemize}

\section{Results}

This section evaluate BERT and Word2Vec across four time spans (3, 5, 10, and 20 years) using four key metrics (SD, MTS, RSC, and LNS). Figure 2 plots the average values with 95\% confidence intervals (±), obtained via bootstrapping, and Appendix 2 reports the full tabulation. Across all time spans, BERT attains substantially higher SD and MTS than Word2Vec, indicating that it preserves more consistent semantic representations both from start to end (SD) and at intermediate intervals (MTS). For instance, over a 3-year window, BERT’s SD is 0.993 ± 0.004 versus 0.790 ± 0.013 for Word2Vec, and its MTS is 0.994 ± 0.003 compared to 0.811 ± 0.011. This pattern holds even at 20 years, where BERT’s SD remains relatively high (0.985 ± 0.005) and MTS is 0.995 ± 0.003; both exceed Word2Vec’s corresponding values (0.746 ± 0.013 and 0.794 ± 0.009). These results suggest that BERT maintains stronger overall semantic stability over both short and long durations, capturing fewer spurious shifts than its static counterpart. 

\textbf{Figure 2. Comparison of Key Metrics Between BERT and Word2Vec Over Different Time Spans}

\begin{figure}[h!]
    \centering
    \includegraphics[width=\textwidth]{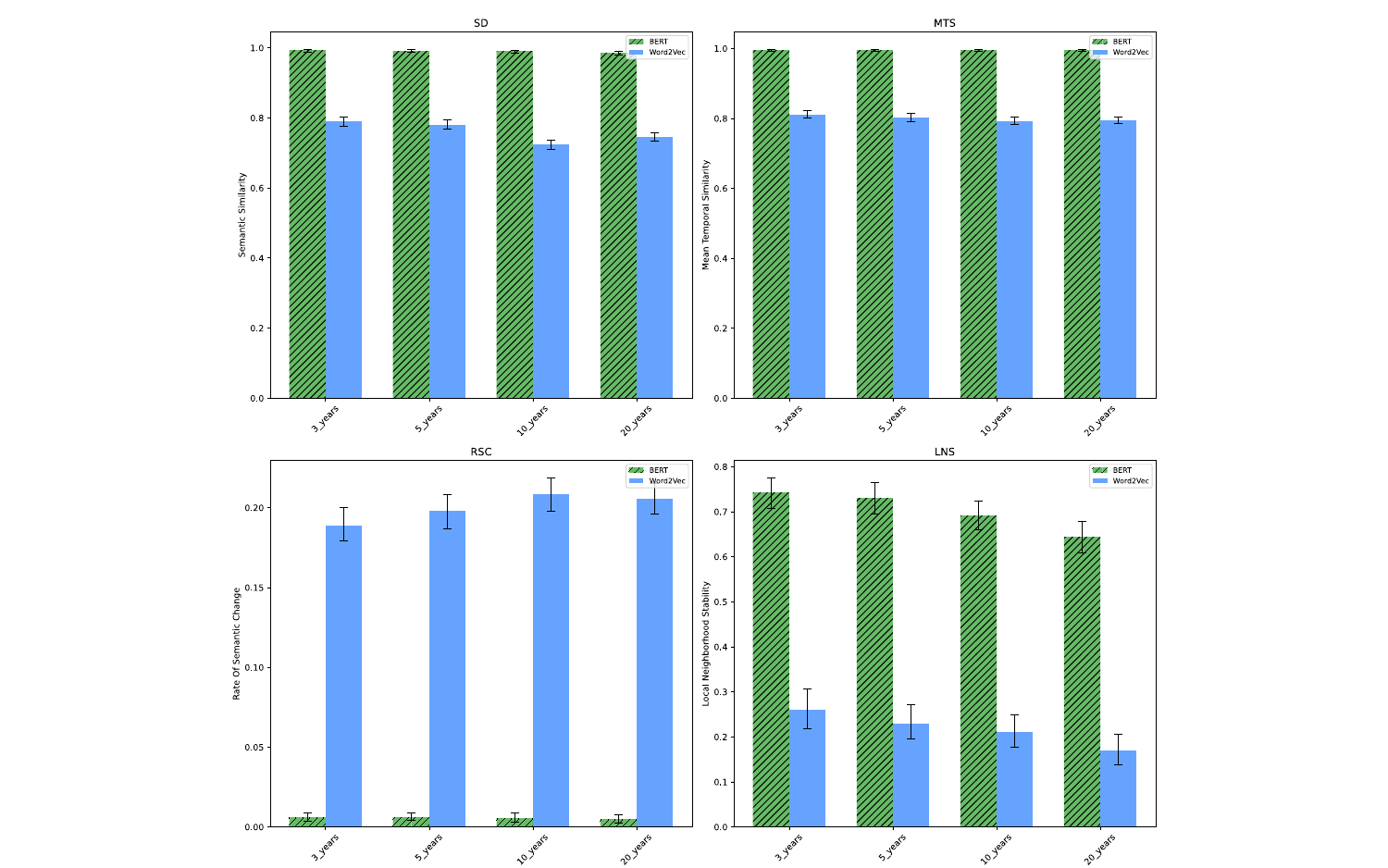} 
\end{figure}

BERT also exhibits markedly lower RSC in every time span. At 3 years, BERT’s RSC is just 0.007 ± 0.002, while Word2Vec’s is 0.189 ± 0.011. Even at 20 years, BERT’s RSC of 0.008 ± 0.002 is far below that of Word2Vec (0.206 ± 0.010). This relatively small rate of change suggests that BERT is able to resist abrupt or erratic drift, enabling it to capture more meaningful long‐term trends without conflating them with ephemeral noise in the data. Combined with the SD and MTS findings, the low RSC highlights BERT’s ability to portray stable embeddings that evolve only when substantial semantic shifts genuinely emerge.

BERT preserves each word’s nearest semantic neighbors more effectively: it consistently achieves higher LNS scores than Word2Vec across all time spans. For instance, in the 3-year window, BERT attains an LNS of 0.741 ± 0.035, nearly three times that of Word2Vec (0.260 ± 0.046). Even at the 20-year mark, BERT’s LNS of 0.644 ± 0.036 remains well above Word2Vec’s 0.170 ± 0.035, suggesting that BERT retains more stable local contexts over both short and extended periods. BERT’s LNS gradually decreases from 0.741 to 0.644 as the analysis window expands from 3 to 20 years. We interpret this modest decline as BERT capturing genuine, long-term semantic evolution (e.g., as certain concepts adopt new connotations or shift in usage) rather than merely “drifting” due to inconsistent training data. By contrast, Word2Vec’s significantly lower LNS values, which also decrease over time, imply that its local neighborhoods undergo larger rearrangements—likely mixing genuine linguistic shifts with model noise. Overall, BERT’s superior LNS aligns with its high SD and MTS scores, reinforcing the model’s capacity to identify meaningful changes in word usage while maintaining a stable representation of each word’s semantic neighborhood.

\section{Discussion and Conclusion}
Taken together, these findings establish that BERT consistently outperforms Word2Vec on core measures of semantic stability (SD, MTS, and RSC) across both short and long time spans, indicating its ability to preserve coherent embeddings when minimal shifts are expected while still adapting to genuine linguistic changes. The elevated LNS values further underscore BERT’s advantage in retaining stable local neighborhoods, even though these neighborhoods gradually evolve over extended periods—a pattern that reflects meaningful semantic development rather than random fluctuations. This balance between stability and sensitivity highlights BERT’s suitability for short‐term discourse analysis as well as longitudinal studies of evolving political or social phenomena.

Figure 3 exemplifies these findings by analyzing the word “diplomacy;” its semantic shifts are visualized using t-SNE bullseye plots for BERT and Word2Vec embeddings. In the Word2Vec plot on the left, the embeddings for “diplomacy” exhibit notable scatter and inconsistencies across years. This dispersion suggests that Word2Vec struggles to maintain stable representations throughout the term. By contrast, the BERT plot reveals a more concentrated and coherent pattern: yearly embeddings form a tighter cluster around the center (representing 2004). The gradual and systematic outward movement of embeddings over time reflects genuine semantic evolution rather than drift, as BERT adapts to changes in the term’s contextual usage.

\textbf{Figure 3. Bullseye Visualization of Yearly Embeddings for ‘Diplomacy’ Using BERT and Word2Vec}

\begin{figure}[h!]
    \centering
    \includegraphics[width=\textwidth]{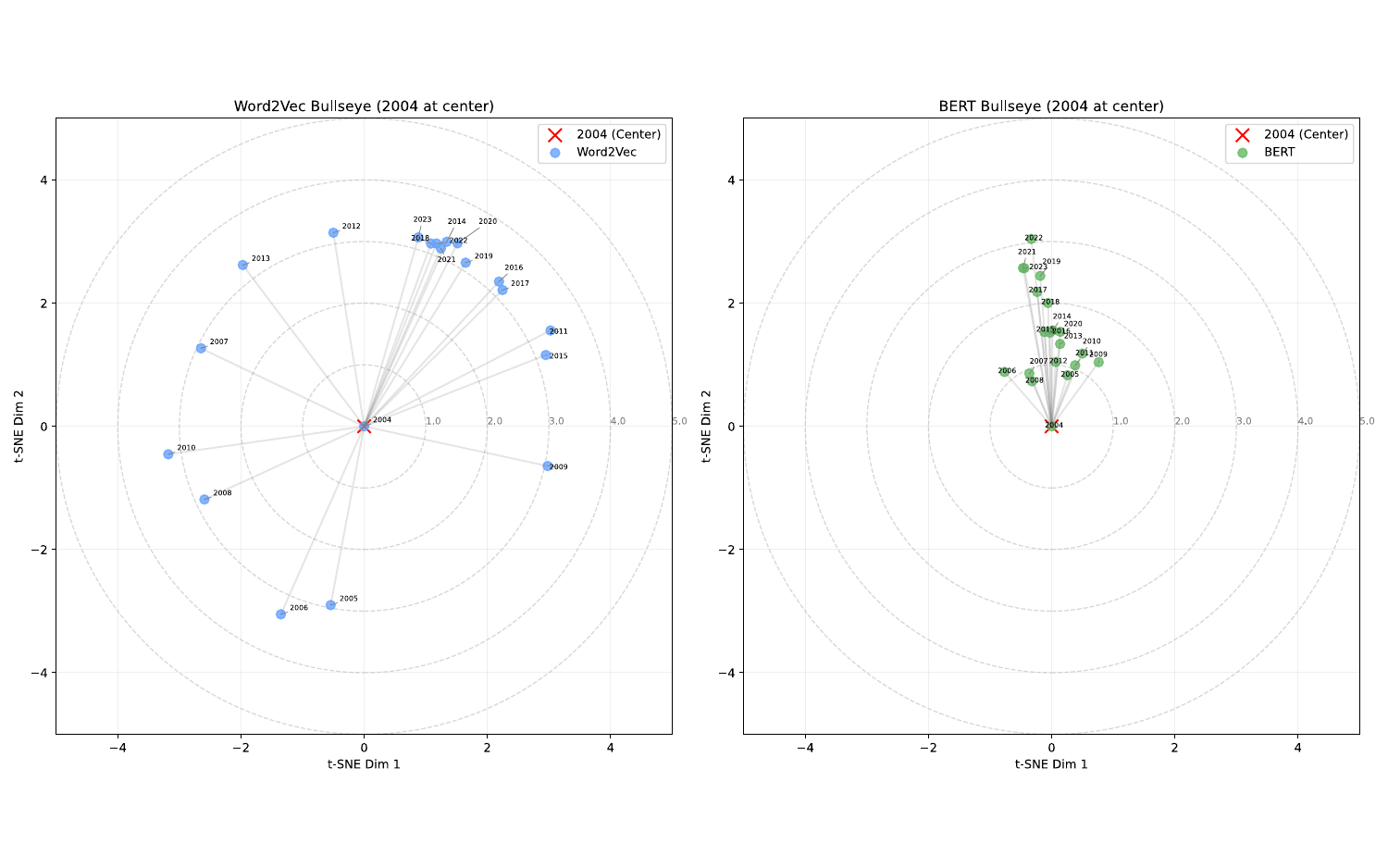} 
\end{figure}

Color is a significant symbol in political science; its meaning often varies across contexts. For example, the term “red” in 

\begin{quote}
\small
…utilizing a great variety of elements of the red gene of our military, and actively adapting to the new situation and tasks of innovation and development… 
\end{quote}

signifies revolutionary ideology, while in 

\vskip 0.2 in

\begin{quote}
\small
…power grid of the province’s municipalities in the next three years of new energy can be accessed to the capacity of the forecast analysis, green, yellow, and red for the warning level labeling… 
\end{quote}

it represents an energy capacity warning. BERT effectively captures this distinction; a cosine similarity of 0.68 between the two contexts reflects their semantic divergence. Word2Vec’s static embeddings fail to differentiate these nuanced meanings, underscoring BERT’s superiority in context-sensitive semantic analysis, which is crucial for understanding complex political and sociocultural narratives.

These findings collectively demonstrate that BERT excels in preserving semantic stability and capturing meaningful shifts in word usage. It outperforms Word2Vec in a variety of linguistic and contextual scenarios. By maintaining stable embeddings over short periods while adapting to genuine semantic evolution across longer intervals, BERT is a powerful tool for analyzing complex, dynamic political discourse. Its ability to distinguish nuanced meanings, as exemplified by the cases of “diplomacy” and “red,” underscores its value for context-sensitive applications in political science, media studies, and policy analysis. This study highlights the limitations of static embeddings like Word2Vec, which often conflate subtle shifts with noise, and showcases how contextual embeddings such as BERT overcome these challenges. The implications of the findings extend beyond the methodological realm: researchers can leverage BERT to trace the evolution of key political concepts, monitor shifts in ideological narratives, and understand the changing connotations of terms in both short- and long-term analyses. Furthermore, its sensitivity to genuine semantic change makes BERT an indispensable resource for investigating sociocultural and political transformations with greater precision.

While BERT demonstrates strong stability in short-term semantic analysis and captures meaningful shifts over longer periods, it is important to acknowledge two limitations. First, the inherent stability of its embeddings, rooted in its pre-training and transformer-based architecture, may occasionally underrepresent the gradual, nuanced semantic drifts that are part of the natural evolution of language. Over extended periods of 30 years or more, this stability could act as a double-edged sword, obscuring subtle cultural or societal changes. This may be particularly linked to the anchoring effect of BERT’s pre-training nature, which, while effective at capturing broad patterns, might not fully reflect the dynamism of language across decades. Second, BERT’s computational demands represent a significant challenge for large-scale analyses. It required hundreds of times more computational resources than Word2Vec in this study. Thus the available resources must be carefully considered when applying BERT to large datasets or longitudinal studies. 

Despite these challenges, this study provides robust evidence of BERT’s superiority over Word2Vec in tasks that require semantic stability and contextual precision, particularly in short- and mid-term social science analyses. BERT’s ability to distinguish nuanced meanings and maintain consistent representations makes it a powerful tool for understanding dynamic political and social phenomena.

In conclusion, while BERT offers a compelling framework for short-term semantic analysis, researchers should remain mindful of its computational costs and potential limitations in capturing gradual semantic shifts. Future advancements in NLP models or hybrid approaches that combine BERT’s stability with methods that are more sensitive to long-term language evolution may provide a more balanced approach to studying the complexities of semantic change over time.

\vskip 0.5 in

\nocite{*} 

\section*{References}
\author{}
\date{}
\maketitle

\begin{spacing}{1}

\begin{hangparas}{\hangindentlength}{1}

\noindent Alrababa’h, A., Marble, W., Mousa, S., \& Siegel, A. A. (2021). Can exposure to celebrities reduce prejudice? The effect of Mohamed Salah on Islamophobic behaviors and attitudes. \textit{American Political Science Review}, \textit{115}(4), 1111–1128.

\noindent Bhardwaj, R., Majumder, N., \& Poria, S. (2021). Investigating gender bias in BERT. \textit{Cognitive Computation}, \textit{13}(5), 1008–1018.

\noindent Bojanowski, P., Grave, E., Joulin, A., \& Mikolov, T. (2017). Enriching word vectors with subword information. \textit{Transactions of the Association for Computational Linguistics}, \textit{5}, 135–146.

\noindent Devlin, J., Chang, M.-W., Lee, K., \& Toutanova, K. (2018). BERT: Pre-training of deep bidirectional transformers for language understanding. \textit{arXiv preprint arXiv:1810.04805}.

\noindent Garg, N., Schiebinger, L., Jurafsky, D., \& Zou, J. (2018). Word embeddings quantify 100 years of gender and ethnic stereotypes. \textit{Proceedings of the National Academy of Sciences}, \textit{115}(16), E3635–E3644.

\noindent Kozlowski, A. C., Taddy, M., \& Evans, J. A. (2019). "The Geometry of Culture: Analyzing the Meanings of Class through Word Embeddings." \textit{American Sociological Review}, \textit{84}(5), 905–949.

\noindent Mikolov, T. (2013). Efficient estimation of word representations in vector space. \textit{arXiv preprint arXiv:1301.3781}.

\noindent Muchlinski, D., Yang, X., Birch, S., Macdonald, C., \& Ounis, I. (2021). We need to go deeper: Measuring electoral violence using convolutional neural networks and social media. \textit{Political Science Research and Methods}, \textit{9}(1), 122–139.

\noindent Pennington, J., Socher, R., \& Manning, C. D. (2014, October). Glove: Global vectors for word representation. In \textit{Proceedings of the 2014 conference on empirical methods in natural language processing (EMNLP)} (pp. 1532–1543).

\noindent Pereira, C., da Silva, R., \& Rosa, C. (2024). How to measure political polarization in text-as-data? A scoping review of computational social science approaches. \textit{Journal of Information Technology \& Politics}, \textit{1}–14.

\noindent Radford, B. J. (2021). Automated dictionary generation for political event coding. \textit{Political Science Research and Methods}, \textit{9}(1), 157–171.

\noindent Rodman, E. (2020). A timely intervention: Tracking the changing meanings of political concepts with word vectors. \textit{Political Analysis}, \textit{28}(1), 87–111.

\noindent Rodriguez, P. L., \& Spirling, A. (2022). Word embeddings: What works, what doesn’t, and how to tell the difference for applied research. \textit{The Journal of Politics}, \textit{84}(1), 101–115.

\noindent Rogers, A., Kovaleva, O., \& Rumshisky, A. (2021). A primer in BERTology: What we know about how BERT works. \textit{Transactions of the Association for Computational Linguistics}, \textit{8}, 842–866.

\noindent Vaswani, A., Shazeer, N., Parmar, N., Uszkoreit, J., Jones, L., Gomez, A. N., Kaiser, Ł., \& Polosukhin, I. (2017). Attention is all you need. \textit{Advances in Neural Information Processing Systems}.

\noindent Widmann, T., \& Wich, M. (2023). Creating and comparing dictionary, word embedding, and transformer-based models to measure discrete emotions in German political text. \textit{Political Analysis}, \textit{31}(4), 626–641.

\noindent Yang, F., Wang, W., Wang, F., Fang, Y., Tang, D., Huang, J., ... \& Yao, J. (2022). scBERT as a large-scale pretrained deep language model for cell type annotation of single-cell RNA-seq data. \textit{Nature Machine Intelligence}, \textit{4}(10), 852–866.

\noindent Yao, Z., Sun, Y., Ding, W., Rao, N., \& Xiong, H. (2018). Dynamic word embeddings for evolving semantic discovery. In \textit{Proceedings of the eleventh ACM international conference on web search and data mining} (pp. 673–681).

\noindent Zhang, Z., Wu, Y., Zhao, H., Li, Z., Zhang, S., Zhou, X., \& Zhou, X. (2020). Semantics-aware BERT for language understanding. In \textit{Proceedings of the AAAI conference on artificial intelligence} (Vol. 34, No. 05, pp. 9628–9635).

\end{hangparas}

\end{spacing}

\newpage

\appendix
\renewcommand{\thesection}{Appendix \arabic{section}.} 

\section{Keywords Table}
\label{appendix_keywords}

\begin{table}[hbt!]
\centering
\begin{threeparttable}
\label{table_keywords}
\setlength{\extrarowheight}{3pt} 
\begin{tabularx}{\textwidth}{|p{0.25\textwidth}|X|}
\hline
\textbf{Category} & \textbf{Keywords} \\
\hline
Political Systems and Government Institutions & Democracy, Constitution, Legislation, Judiciary, Political Parties, Politics, Bureaucracy, Administration, Power, Rule of Law, State, Law, Elections, Voting, Public, Policies, Parliament, President, Prime Minister \\
\hline
Political Ideologies & Republic, Sovereign, Authoritarianism, Socialism, Liberal, Conservative, Radical, Nationalism, Environmental, Progressive \\
\hline
International Relations and Globalization & Diplomacy, International, Coalition, Global, Trade, War, Peace, Conflict, Allies, Sanctions \\
\hline
Political Behavior and Society & Voters, Citizenship, Trust, Participation, Mobilization, Dissemination, Polarization, Comparison, Monitor, Protest \\
\hline
Public Administration and Political Economy & Management, Efficiency, Reform, Services, Budget, Fiscal, Economy, Taxes, Investments, Welfare \\
\hline
\end{tabularx}
\end{threeparttable}
\end{table}

\section{Detailed Results of Comparative Analysis Between BERT and Word2Vec Over Different Time Spans}

This appendix provides the detailed quantitative results of the metrics displayed in Figure 2. The table below presents the metrics for each time span, including Semantic Displacement (SD), Mean Temporal Similarity (MTS), Rate of Semantic Change (RSC), and Local Neighborhood Stability (LNS), and for both BERT and Word2Vec models. The 95\% confidence intervals are included alongside each value as error margins, showing the variability in the data due to sampling effects.

\begin{table}[hbt!]
\centering
\begin{threeparttable}
\label{table_model_comparison}
\setlength{\extrarowheight}{3pt} 
\begin{tabularx}{\textwidth}{
    p{0.12\textwidth} 
    p{0.1\textwidth} 
    >{\centering\arraybackslash}p{0.15\textwidth} 
    >{\centering\arraybackslash}p{0.15\textwidth} 
    >{\centering\arraybackslash}p{0.15\textwidth} 
    >{\centering\arraybackslash}p{0.15\textwidth} 
}
\toprule
\textbf{Time Span} & \textbf{Model} & \textbf{SD} & \textbf{MTS} & \textbf{RSC} & \textbf{LNS} \\
\midrule
3 years & BERT & $0.993 \pm 0.004$ & $0.994 \pm 0.003$ & $0.006 \pm 0.003$ & $0.741 \pm 0.035$ \\
3 years & W2V  & $0.790 \pm 0.013$ & $0.811 \pm 0.011$ & $0.189 \pm 0.011$ & $0.260 \pm 0.046$ \\
\midrule
5 years & BERT & $0.991 \pm 0.005$ & $0.994 \pm 0.003$ & $0.006 \pm 0.003$ & $0.729 \pm 0.035$ \\
5 years & W2V  & $0.781 \pm 0.013$ & $0.802 \pm 0.011$ & $0.198 \pm 0.011$ & $0.230 \pm 0.041$ \\
\midrule
10 years & BERT & $0.990 \pm 0.004$ & $0.995 \pm 0.003$ & $0.005 \pm 0.003$ & $0.691 \pm 0.033$ \\
10 years & W2V  & $0.724 \pm 0.013$ & $0.792 \pm 0.011$ & $0.208 \pm 0.011$ & $0.211 \pm 0.038$ \\
\midrule
20 years & BERT & $0.985 \pm 0.005$ & $0.995 \pm 0.003$ & $0.005 \pm 0.003$ & $0.644 \pm 0.036$ \\
20 years & W2V  & $0.746 \pm 0.013$ & $0.794 \pm 0.009$ & $0.206 \pm 0.010$ & $0.170 \pm 0.035$ \\
\bottomrule
\end{tabularx}
\end{threeparttable}
\end{table}

\end{document}